\documentclass[11pt]{article}

\usepackage[final]{acl}

\usepackage{times}
\usepackage{latexsym}

\usepackage[T1]{fontenc}

\usepackage[utf8]{inputenc}

\usepackage{microtype}

\usepackage{inconsolata}

\usepackage{comment}

\usepackage{graphicx}
\usepackage{amsmath}
\usepackage{multirow}
\usepackage{subcaption}
\usepackage{enumitem}

\title{From Recognition to Reasoning: Benchmarking and Enhancing MLLMs on Real-World Receipt Document Understanding}

\author{
  Yandi Wang\thanks{\ \ Equal contribution.} , 
  Libin Zhan\footnotemark[1] ,
  Ziwei Huang\footnotemark[1], 
  Tiancheng Luo, \\ 
  \textbf{Yuxuan Jiang, Wang Dong, 
  Leilei Gan\thanks{\ \ Corresponding authors.}, 
  Jun Chen\footnotemark[2]}
  \\
  Zhejiang University, China \\
  \texttt{\{yandiwang, zhanlibin, leileigan, chenjun332\}@zju.edu.cn}
}

\newcommand{\datasetname}{ReceiptBench} 

\begin{document}
\maketitle

\begin{abstract}

Extracting structured information from visual documents (Visual Information Extraction, VIE) is a cornerstone of business automation. While recent Multimodal Large Language Models (MLLMs) have shown promising capabilities, existing benchmarks suffer from critical limitations in scale and realism, lack semantic granularity, and fail to cover diverse document types. To bridge this gap, we introduce \textbf{\datasetname}, a large-scale, human-annotated benchmark consisting of 10k diverse receipts, organizing information extraction into four hierarchical sub-tasks: (1) \textit{Basic Perception} for raw text spotting, (2) \textit{Format Normalization} for strictly following standardization instructions, (3) \textit{Semantic Reasoning} for inferring implicit attributes from context, and (4) \textit{Structure Parsing} for handling nested line items. Furthermore, we propose a two-stage training framework incorporating \textit{Metric-Aware Group Relative Policy Optimization (GRPO)}, which translates rigorous evaluation constraints into reinforcement learning signals to enhance structural consistency. Extensive experiments demonstrate that our method yields state-of-the-art performance, surpassing leading proprietary models on complex reasoning tasks. We release our datasets and code at \url{https://github.com/wwwT0ri/ReceiptBench}.
\end{abstract}

\begin{figure*}[t] 
    \setlength{\abovecaptionskip}{0.2cm}
    \setlength{\belowcaptionskip}{-0.3cm}
    \centering
    \includegraphics[width=\textwidth]{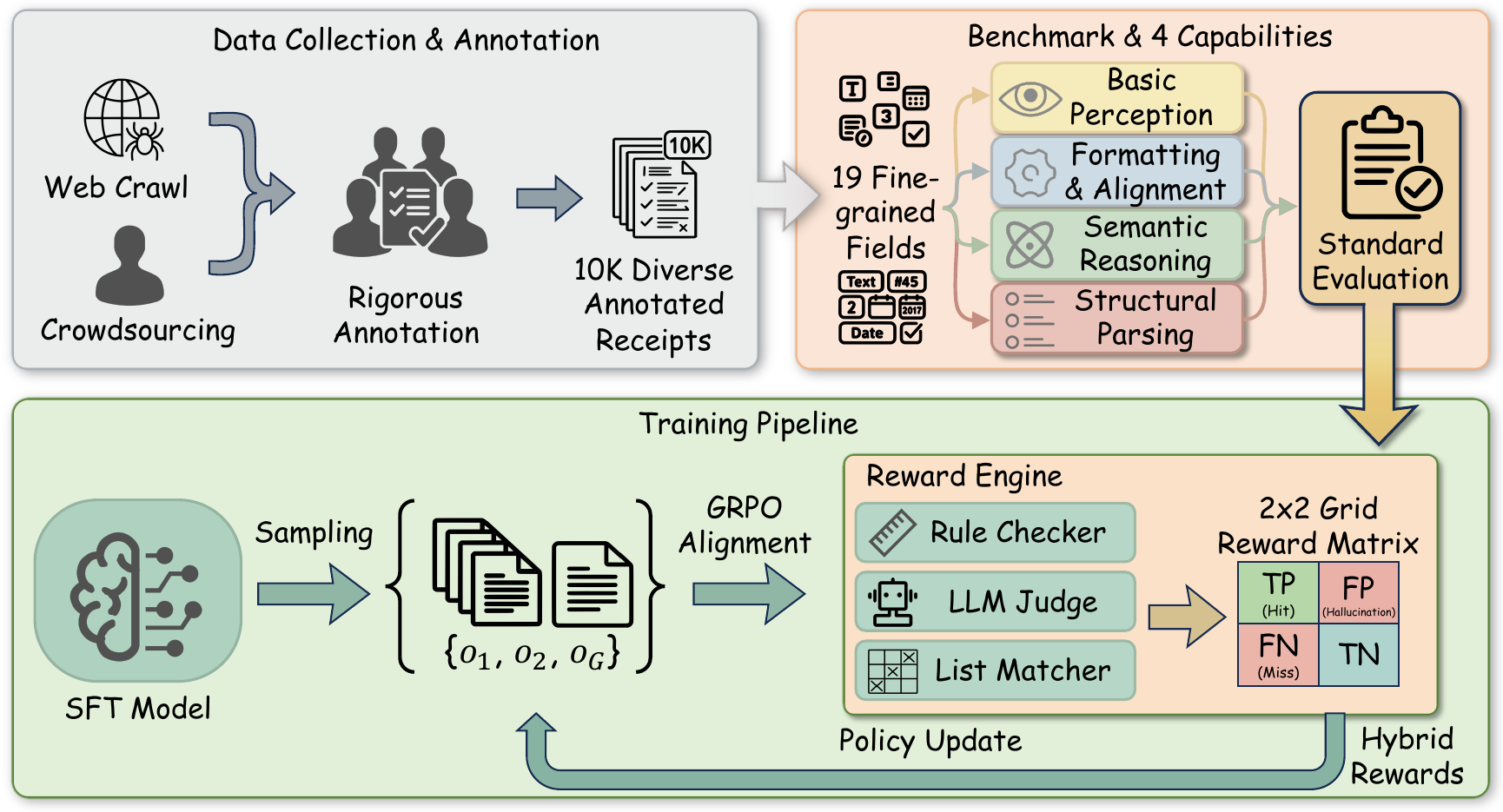} 
    
    \caption{\textbf{Overview of the \datasetname\ Framework.} 
    \textbf{(Top) Benchmark Construction:} We curate 10k diverse invoices via web crawling and crowdsourcing. The benchmark defines a hierarchical taxonomy covering four capabilities: \textit{Basic Perception}, \textit{Formatting}, \textit{Semantic Reasoning}, and \textit{Structural Parsing}.
    \textbf{(Bottom) Training Pipeline:} To master these capabilities, we propose a Metric-Aware GRPO framework. The SFT model acts as the policy, generating outputs that are evaluated by a hybrid Reward Engine (comprising Rule Checkers, LLM Judges, and List Matchers). Crucially, the evaluation results are mapped into a 2x2 Reward Matrix—rewarding hits (TP) while explicitly penalizing hallucinations (FP)—to align the model with rigorous auditing standards.}
    \label{fig:main_architecture}
    \vspace{-3mm}
  \end{figure*}

\section{Introduction}
Visual Information Extraction (VIE) serves as a cornerstone of enterprise automation, enabling the digitization of workflows in finance, logistics, and legal domains. 
The recent emergence of Multimodal Large Language Models (MLLMs) \citep{bai2023qwen, hurst2024gpt, chen2024internvl} has shifted the paradigm from pipeline-based Optical Character Recognition (OCR) to end-to-end visual reasoning.

The advancement of VIE, particularly for Key Information Extraction (KIE), relies heavily on high-quality benchmarks to ensure the extraction reliability and logical consistency required for financial evidence. Receipts and invoices are critical in this domain due to their global ubiquity and layout diversity. However, as highlighted in Table~\ref{tab:comparison}, existing benchmarks struggle to simultaneously satisfy the demands of scale, diversity, and granularity. Early real-world datasets~\cite{huang2019icdar, park2019cord, sun2021spatial, wang2021towards, xu2022xfund} established essential baselines but are severely limited in scale (<2k images) and confined to narrow domains (e.g., retail and dining receipts), failing to represent the heterogeneous layouts found in broader real-world scenarios. While synthetic datasets like FATURA~\citep{limam2025fatura} address the data volume issue, they often rely on finite templates and lack the authentic semantic logic inherent in real transactions. Furthermore, current efforts~\citep{abdallah2024receiptsense, huang2019icdar, park2019cord, mathew2021docvqa, jaume2019funsd, xu2022xfund} predominantly focus on explicit text extraction; this shallow perception fails to capture the complexity of real-world financial processing, which demands format normalization, implicit reasoning, and structural parsing.

To bridge this gap, we introduce \textbf{\datasetname}, a large-scale benchmark designed to evaluate reasoning-aware information extraction from complex financial documents. 
\datasetname\ comprises \textbf{10,656} high-quality images collected from diverse real-world sources, covering multi-lingual regions and heterogeneous document types (e.g., taxi invoices, ferry tickets, hotel statements). 
Unlike previous works that treat extraction as a flat slot-filling task, we propose a hierarchical taxonomy of four capabilities: \textit{Perception}, \textit{Normalization}, \textit{Reasoning}, and \textit{Structure}.
This taxonomy requires models to not only "read" the pixels but also "understand" the business logic and "structure" the output rigorously.


However, our evaluation on ReceiptBench highlights significant deficiencies in current methodologies: general-purpose models~\citep{bai2023qwen, hurst2024gpt, chen2024internvl} often overlook fine-grained financial constraints, while specialized models~\citep{chen2025dianjin, cui2025paddleocr, huang2022layoutlmv3} lack the generative reasoning required for complex extraction. Even establishing a competitive baseline via standard Supervised Fine-Tuning (SFT) proves non-trivial. SFT optimizes for local token probabilities rather than global logical consistency, frequently resulting in structural hallucinations (e.g., invalid JSON syntax) and arithmetic inconsistencies (e.g., line items not summing to the total). To address this, we propose a two-stage training framework. After initial instruction tuning, we introduce an alignment stage using Group Relative Policy Optimization (GRPO)~\citep{shao2024deepseekmath}. Specifically, we design a \textbf{Metric-Aware Reward Engine} that directly translates our rigorous evaluation protocols—arithmetic checks and schema adherence—into reinforcement learning signals. This approach explicitly penalizes hallucinations and rewards logical coherence, enabling the model to internalize the complex reasoning rules of the benchmark.

In summary, our contributions are as follows:
\begin{itemize}
    \item We present \textbf{\datasetname}, a challenging benchmark with 10k real-world samples and 19 fine-grained fields, shifting the focus of VIE from literal extraction to cognitive reasoning and structural parsing.
    \item We design a robust \textbf{Hybrid Evaluation Protocol} that moves beyond simple string matching, incorporating LLM-based semantic judges and Hungarian matching algorithms~\citep{kuhn1955hungarian} for nested lists to ensure fair and accurate assessment.
    \item We propose a \textbf{Metric-Aware GRPO} training framework. Extensive experiments demonstrate that this method significantly improves the reasoning and structural capabilities of open-source MLLMs (e.g., Qwen3-VL~\citep{bai2025qwen3}), narrowing the gap with proprietary SOTA models like GPT-5.
\end{itemize}

\section{Related Work}


\subsection{Benchmarks for VIE}
Existing benchmarks face critical limitations in \textbf{scale and realism}. Early datasets like SROIE~\citep{huang2019icdar}, CORD~\citep{park2019cord}, and WildReceipt~\citep{sun2021spatial} are too small ($<2$k images) for data-hungry MLLMs. Comprehensive benchmarks like CC-OCR~\citep{yang2025cc} offer limited fresh KIE challenges by partially aggregating existing datasets ($\sim$2k samples). While FATURA~\citep{limam2025fatura} addresses scalability via synthesis, it suffers from template bias and lacks authentic semantic logic.

Regarding \textbf{granularity and task alignment}, existing efforts often diverge from the needs of enterprise automation. ReceiptSense~\citep{abdallah2024receiptsense} provides sparse annotations that hinder complex reasoning. Meanwhile, benchmarks like DocVQA~\citep{mathew2021docvqa}, MP-DocVQA~\citep{tito2023hierarchical}, and DUDE~\citep{van2023document} frame document understanding primarily as open-ended Visual Question Answering (VQA) or generic layout analysis (e.g., FUNSD~\citep{jaume2019funsd}, XFUND~\citep{xu2022xfund}) rather than structured schema-constrained extraction. Furthermore, while OCR-Reasoning~\citep{huang2025ocr} extensively evaluates visual reasoning, its taxonomy is heavily skewed toward academic problem-solving rather than the strict financial business logic required in real-world scenarios.

Finally, \textbf{document diversity} remains narrow; datasets are often skewed towards simple retail slips (SROIE) or exam papers (EPHOIE~\citep{wang2021towards}), failing to represent heterogeneous financial documents like multi-page hotel statements. Consequently, the field lacks a unified benchmark balancing these critical dimensions, a gap \textbf{\datasetname} aims to fill.

\subsection{Specialized Document Understanding Models}
Early pipeline systems combined OCR with NLP models. While the LayoutLM series~\citep{xu2020layoutlm, huang2022layoutlmv3} embedded spatial semantics, they still \textbf{relied on external OCR}. Subsequently, Donut~\citep{kim2022ocr} and Nougat~\citep{blecher2023nougat} introduced OCR-free, end-to-end paradigms mapping pixels to text. Recent models prioritize efficiency: GOT-OCR~\citep{wei2024general} unifies OCR tasks under a general theory, while PaddleOCR-VL~\citep{cui2025paddleocr} utilizes a NaViT-style encoder to achieve SOTA performance with minimal consumption.

To address high-resolution token costs, DeepSeek-OCR~\citep{wei2025deepseek} introduces optical context compression to minimize vision tokens. Similarly, UReader~\citep{ye2023ureader} and mPLUG-DocOwl 1.5~\citep{hu2024mplug} employ shape-adaptive cropping. Moving towards agentic reasoning, DianJin-OCR~\citep{chen2025dianjin} leverages Chain-of-Thought (CoT)~\citep{wei2022chain} for interleaved planning and tool use. However, current benchmarks lack evaluation for such \textbf{complex reasoning and structural parsing}, motivating \textbf{\datasetname}.

\begin{table*}[t]
\setlength{\abovecaptionskip}{0.2cm}
\setlength{\belowcaptionskip}{-0.3cm}
\centering
\small
\resizebox{\textwidth}{!}{%
\begin{tabular}{l|c|c|l|l}
\hline
\textbf{Dataset} & \textbf{Size} & \textbf{Fields} & \textbf{Document Types Coverage} & \textbf{Key Limitations} \\ \hline
SROIE~\cite{huang2019icdar} & 1,000 & 4 & Retail Receipts & Small Scale; Low Granularity \\
FUNSD~\cite{jaume2019funsd} & 199 & - & General Forms & Small Scale; Generic Entity Labels \\
CORD~\cite{park2019cord} & 1,000 & 8 & Retail \& Dining Receipts & Small Scale; Narrow Domain \\
WildReceipt~\cite{sun2021spatial} & 1,765 & 25 & Retail Receipts & Small Scale; Narrow Domain \\
EPHOIE~\cite{wang2021towards} & 1,494 & 10 & Examination Papers & Small Scale; Education Domain \\
XFUND~\cite{xu2022xfund} & 1,393 & - & General Forms & Small Scale; Generic Entity Labels\\ \hline
DocVQA~\cite{mathew2021docvqa} & 12,767 & - & Various Documents & Paradigm Divergence (QA vs. IE) \\
FATURA~\cite{limam2025fatura} & 10,000 & 24 & Invoices & Synthetic Logic; Finite Templates (50) \\
ReceiptSense~\cite{abdallah2024receiptsense} & 20,000 & 5 & Retail Receipts & Low Granularity; Narrow Domain \\ \hline
\textbf{ReceiptBench (Ours)} & \textbf{10,656} & \textbf{19} & \textbf{Purchasing, Hotel, Travel, etc.} & \textbf{-} \\ \hline
\end{tabular}%
}
\caption{\textbf{Comparison of our dataset with existing VIE benchmarks.} Existing datasets exhibit critical limitations in three key aspects: (1) \textbf{Scale and Realism}: they are either limited in size or rely on synthetic generation; (2) \textbf{Granularity and Task Alignment}: they suffer from low granularity or target divergent paradigms such as QA and generic layout analysis; and (3) \textbf{Document Diversity}: they are restricted to specific narrow domains like retail or non-financial domains. In contrast, our dataset balances scale, realism, granularity, and document diversity.}
\vspace{-3mm}
\label{tab:comparison}
\end{table*}

\section{The \datasetname\ Benchmark}
\label{sec:dataset}

To address the limitations of existing benchmarks and support the training of end-to-end MLLMs for complex document understanding, we introduce \textbf{\datasetname}, a large-scale, real-world dataset designed with financial accounting standards and multi-dimensional capability evaluation in mind.

\subsection{Data Collection and Annotation}

\paragraph{Data Sources.} 
The dataset comprises \textbf{10,656} images sourced from real-world scenarios to ensure diversity in layout, visual conditions, and content. Our collection followed a hybrid strategy: (1) \textit{Public Web Crawling:} We gathered receipt images from publicly available repositories, prioritizing those with varied layouts and quality. (2) \textit{Crowdsourced Solicitation:} To capture long-tail document types (e.g., specific regional taxi invoices or flight tickets) often absent in public collections, we conducted a paid, questionnaire-driven campaign, ensuring broad geographical and domain coverage. Each collected document was manually reviewed to verify its authenticity and legibility.

\paragraph{Annotation Process.}
We engaged a professional data annotation service to ensure high-quality labeling. The entire dataset was divided into 10 batches, each of which underwent the vendor's internal annotation and multi-stage review cycle. Upon delivery, we applied a stringent \textbf{acceptance protocol} comprising three validation stages:

\textbf{1. Random Sampling Inspection:} We randomly sampled a validation subset, in which domain experts manually verified the correctness of all annotated fields to ensure accuracy.

\textbf{2. Automated Logic and Format Validation:} We employed custom validation scripts to check compliance with the predefined schema. This included crucial \textbf{cross-field consistency} checks (e.g., verifying that the sum of line items in \texttt{detail} matches \texttt{std\_total}) and \textbf{standard formatting} rules (e.g., ensuring \texttt{std\_invoice\_time} conforms to the standard date pattern).

\textbf{3. Error Analysis and Iterative Refinement:} Validation results were aggregated to compute field-specific accuracy rates and to summarize common error patterns. These findings, encompassing all detected errors, were documented in an audit report provided to the vendor. If the accuracy for any field within the validation subset fell below the \textbf{97\%} threshold, the \textit{entire} batch was rejected and required to undergo revision and re-annotation.

Through this multi-stage, iterative pipeline, the final dataset achieved an overall average annotation accuracy of \textbf{98.7\%}, confirming its high quality for subsequent tasks.

\subsection{Dataset Statistics}

\textbf{Document Type Diversity.} 
\datasetname\ distinguishes itself by covering a wide spectrum of service-oriented financial documents, moving beyond the retail-centric distribution of prior works (Table~\ref{tab:type_distribution}).
While \textit{General Purchase \& Dining} receipts account for the plurality (43.5\%), the dataset features a substantial proportion of \textit{Transportation} documents (Plane, Taxi, Train, Bus, etc.), totaling over 35\%, as well as complex \textit{Hotel Bills} (11.2\%). This distribution introduces multi-page layouts and tabular structures significantly more challenging than standard supermarket receipts.

\begin{table}[h]
\setlength{\abovecaptionskip}{0.1cm} 
\setlength{\belowcaptionskip}{-0.3cm} 
\centering
\footnotesize 
\renewcommand{\arraystretch}{0.95} 
\setlength{\tabcolsep}{4pt} 
\begin{tabular}{l|r|r}
\hline
\textbf{Category} & \textbf{Count} & \textbf{Pct.} \\ \hline
Purchase \& Dining & 4,636 & 43.50\% \\
Plane Ticket & 2,175 & 20.42\% \\
Hotel Bill & 1,198 & 11.24\% \\
Taxi Receipt & 1,132 & 10.62\% \\
Train Ticket & 416 & 3.90\% \\
Bus Ticket & 249 & 2.34\% \\
Ship/Ferry Ticket & 150 & 1.41\% \\
Fuel Receipt & 132 & 1.24\% \\
Metro Ticket & 130 & 1.22\% \\ \hline
Others & 438 & 4.11\% \\ \hline
\textbf{Total} & \textbf{10,656} & \textbf{100.0\%} \\ \hline
\end{tabular}
\caption{Distribution of document types in our dataset. "Others" includes Car Rental, Postage, Toll, Parking, Internet, Phone, Baggage, Water, Electricity, Medical, Education and Handling receipts.}
\vspace{-3mm}
\label{tab:type_distribution}
\end{table}

\paragraph{Language Distribution.} 
As shown in Table~\ref{tab:lang_distribution}, the dataset is predominantly English (98.0\%) to align with the primary pre-training data of most MLLMs. However, it includes a "long-tail" of 213 samples covering 8 other languages. This inclusion allows for evaluating the model's robustness against linguistic noise and character variations in low-resource scenarios.

\subsection{Task Taxonomy and Schema}
\label{sec:schema}


We define the information extraction problem as a set of four progressive sub-tasks targeting 19 distinct fields (e.g., \texttt{std\_invoice\_time, tax\_number}). The selection of these fields is grounded in standard accounting principles, ensuring the benchmark's utility for real-world financial auditing. See Appendix~\ref{annoSchema} for detailed definitions.

Based on the cognitive capabilities required, we partition the fields into four sub-tasks:

\paragraph{Task 1: Basic Perception (8 fields).} Evaluates Optical Character Recognition (OCR) and grounding. It targets explicit text such as \texttt{invoice\_number} and raw timestamps. Success here indicates the model can accurately "read" visual tokens \citep{biten2019scene}.
    
\paragraph{Task 2: Formatting \& Normalization (4 fields).} Tests instruction-following abilities. Models must convert raw text into standardized formats (e.g., converting "20 Oct, 23" to "2023-10-20" for \texttt{std\_start\_time}). This aligns with the instruction tuning paradigm critical for LLM usability \citep{wei2021finetuned}.
    
\paragraph{Task 3: Semantic Reasoning (6 fields).} Requires extracting implicit information. For instance, deducing \texttt{type}="Hotel" from room charges, or inferring \texttt{std\_curr}="USD" from a "New York" address. This evaluates multi-modal reasoning beyond simple extraction \citep{xu2020layoutlm}.
    
\paragraph{Task 4: Structural Parsing (1 field).} The \texttt{detail} field requires parsing complex, often nested tables into a list of dictionaries (content, amount, tax status). This represents the most challenging task, demanding an understanding of spatial structures similar to table extraction benchmarks \citep{zhong2020publaynet}.

\subsection{Evaluation Protocol}
\label{sec:eval_protocol}

Evaluating information extraction from complex invoices presents unique challenges, such as valid OCR variations and permutation-invariant lists. To ensure a robust and fair comparison for \datasetname, we define a standardized hybrid evaluation protocol combining rule-based matching and semantic judgment.

\paragraph{Hierarchical Evaluation Logic.}
We categorize the 19 target fields into four types, applying specific metrics for each:

\textbf{Type A: Exact Match Fields.} For fields requiring strict adherence to visual evidence (e.g., \texttt{type}, \texttt{tax\_number}, \texttt{std\_invoice\_time}), we use \textbf{Exact Match (EM)}. Both ground truth and predictions are normalized (lowercased, whitespace trimmed) before comparison to handle minor spacing differences.
    
\textbf{Type B: Numeric Fields.} For monetary values (e.g., \texttt{std\_total}), we allow a floating-point tolerance of $\epsilon < 1e^{-6}$. Zero values (0) and empty strings are treated equivalently to handle format inconsistencies.
    
\textbf{Type C: Semantic Fields.} For fields where minor textual variations preserve meaning (e.g., \texttt{place}, \texttt{seller\_name}), we employ a \textbf{Cascading Judge}: 
(1) \textit{Exact Filter:} First, we check for normalized string equality. If they match, it is counted as a True Positive (TP).     
(2) \textit{LLM Judge:} If the exact match fails, we employ a lightweight LLM (Qwen3-4B) as a semantic judge. The model is prompted with specific criteria \textbf{(see Table \ref{tab:judge_prompt})} to determine if the predicted entity is semantically equivalent to the ground truth, explicitly allowing for abbreviations (e.g., "Co." vs. "Company") and synonyms while penalizing factual errors.

\textbf{Type D: Structured List Fields.} Evaluating the lists (e.g., \texttt{detail}, \texttt{orig\_curr}) is the most challenging aspect due to order invariance and nested attributes. We formulate this as a \textbf{Maximum Bipartite Matching} problem. 
    For a predicted list $P$ and ground truth list $G$, we construct a cost matrix $C$ where $C_{ij}$ represents the dissimilarity between item $P_i$ and $G_j$. 
    The dissimilarity is derived from a composite similarity score $S_{ij}$, calculated as a weighted sum of four metrics to capture both lexical and semantic correspondence:
    \begin{equation}
        S_{ij} = \alpha \cdot S_{\text{lev}} + \beta \cdot S_{\text{sort}} + \gamma \cdot S_{\text{lcs}} + \delta \cdot S_{\text{sem}}
    \end{equation}
    where $S_{\text{lev}}$ denotes the Levenshtein ratio, $S_{\text{sort}}$ is the Token Sort similarity (robust to word reordering), $S_{\text{lcs}}$ is the Longest Common Subsequence ratio, and $S_{\text{sem}}$ represents the cosine similarity of embeddings from a Sentence Transformer. 
    The coefficients $\alpha, \beta, \gamma, \delta$ are hyperparameters empirically optimized via grid search on a human-annotated validation set to maximize alignment with human judgment. A comprehensive hyperparameter sensitivity analysis (detailed in Appendix \ref{app:sensitivity}) confirms that our evaluation metric and the resulting model rankings are highly robust to these weight variations.
    
    The cost is defined as $C_{ij} = 1 - S_{ij}$. For the \texttt{detail} field, we impose additional hard constraints: items with mismatched numerical amounts ($|\Delta| > 0.05$) are assigned an infinite cost. We then apply the \textbf{Hungarian Algorithm} to find the optimal assignment, accepting matches only if the cost $C_{ij} \le 0.25$ and attributes align.

\paragraph{Primary Metrics.}
We mainly report \textbf{F1-score}. 
Given that many fields in \datasetname\ can be legitimately empty (e.g., \texttt{departure} for a restaurant receipt), correct identification of absent information is crucial. Therefore, our evaluation script explicitly accounts for True Negatives (TN) to avoid penalizing models that correctly predict "empty" for missing fields.

\section{Methodology}
\label{sec:method}

While one-shot evaluation on proprietary models (e.g., GPT-5) provides a reference for upper-bound performance, it is crucial to establish strong open-source baselines to validate the learnability of \datasetname. In this section, we describe our two-stage training pipeline: Supervised Fine-Tuning (SFT) for instruction adherence and Group Relative Policy Optimization (GRPO) \citep{shao2024deepseekmath} for reasoning alignment.

\subsection{Stage 1: Supervised Fine-Tuning (SFT)}
To equip the model with the capability to handle the complex extraction rules of \datasetname, we construct a rigorous instruction-following dataset.

\paragraph{Instruction Schema Design.}
Unlike general captioning tasks, our task requires strict adherence to a predefined JSON schema. We designed a comprehensive system prompt (see Appendix \ref{app:prompts} for full text) that includes:
(1) \textbf{Role Definition:} An AI assistant specialized in invoice processing.
(2) \textbf{Field Constraints:} Explicit rules for 19 fields (e.g., \texttt{type} must be chosen from a fixed list of 8 categories; \texttt{std\_total} must be rounded to 2 decimal places).
(3) \textbf{Negative Constraints:} Instructions on how to handle missing fields (return empty string `""' rather than `null').

Formally, for each image $I$, we construct the instruction prompt $P$ and the ground truth JSON $Y$. The SFT objective is to minimize the negative log-likelihood of the output tokens given the image and instruction.

\subsection{Stage 2: Alignment via GRPO}
\label{sec:grpo}

While SFT instills the basic instruction-following capabilities, models often struggle with the precise trade-off between extraction recall and hallucination suppression. To align the model's behavior with the strict standards of \datasetname, we employ Group Relative Policy Optimization (GRPO) \citep{shao2024deepseekmath}.

\paragraph{Metric-Aware Reward Shaping.}
Unlike generic RLHF which relies on a separate reward model, we construct a rule-based reward function directly derived from our evaluation protocol (Section \ref{sec:eval_protocol}).
For each field $f$ in the invoice, let $P_f$ be the predicted value and $G_f$ be the ground truth. We define the similarity score $S(P_f, G_f) \in [0, 1]$ based on the field type (e.g., Exact Match score, or the LLM-Judge score for semantic fields, or the Hungarian matching score for lists).

To handle the sparsity of invoice fields, we introduce a \textbf{Reward Shaping} mechanism based on the confusion matrix states (TP, TN, FP, FN). The reward $R_f$ for field $f$ is defined as follows:

\vspace{-0.5\baselineskip}
\begin{equation}
    \renewcommand{\arraystretch}{0.8} 
    R_f(P_f, G_f) = 
    \begin{cases} 
    S(P_f, G_f) & \text{if } G_f \neq \emptyset \land P_f \neq \emptyset \quad \\
    \lambda_{\text{TN}} & \text{if } G_f = \emptyset \land P_f = \emptyset \quad \\
    \lambda_{\text{FP}} & \text{if } G_f = \emptyset \land P_f \neq \emptyset \quad \\
    \lambda_{\text{FN}} & \text{if } G_f \neq \emptyset \land P_f = \emptyset \quad
    \end{cases}
\end{equation}
\vspace{-0.5\baselineskip}

\noindent where the hyperparameters are set as follows:

\textbf{True Positive (TP):} The reward is the alignment score $S \in [0, 1]$. For semantic fields, this $S$ is provided by the LLM Judge (Section \ref{sec:eval_protocol}), encouraging semantically correct answers even if they aren't exact string matches.

\textbf{True Negative (TN, $\lambda_{\text{TN}} = 0.3$):} We assign a modest positive reward. This encourages the model to correctly identify missing fields but prevents "mode collapse" where the model learns to maximize rewards by simply outputting empty strings for everything (which would happen if $\lambda_{\text{TN}}$ were too high).

\textbf{False Positive (FP, $\lambda_{\text{FP}} = -0.5$):} We impose a negative penalty to explicitly suppress hallucinations, which is critical for financial document processing.

\textbf{False Negative (FN, $\lambda_{\text{FN}} = 0$):} No reward is given when the model fails to extract existing information.

The final reward for an invoice is the average of rewards across all 19 fields. By optimizing this shaped reward, GRPO effectively fine-tunes the model's decision boundary between "answering" and "abstaining."

\begin{table*}[t]
\setlength{\abovecaptionskip}{0.2cm}
\setlength{\belowcaptionskip}{-0.3cm}
\centering
\small
\resizebox{\textwidth}{!}{
\begin{tabular}{l|l|c|c|cccc}
\hline
\textbf{Category} & \textbf{Model} & \textbf{Size} & \textbf{Overall} & \textbf{Perception} & \textbf{Normalization} & \textbf{Reasoning} & \textbf{Structure} \\
\hline
\multirow{4}{*}{\shortstack[l]{Proprietary\\\& API-based}} 
& GPT-5 & - & 0.7076 & 0.7304 & 0.8743 & 0.8706 & 0.4893 \\
& Gemini-3-Pro & - & 0.7373 & 0.7360 & 0.9086 & 0.8714 & 0.5781 \\
& InternVL3.5-241B & 241B & 0.6742 & 0.6853 & 0.8791 & 0.8024 & 0.5112 \\
& Qwen3-VL-Plus & - & 0.7210 & 0.7306 & 0.9000 & \textbf{0.8787} & 0.5484 \\
\hline
\multirow{6}{*}{\shortstack[l]{General Open}} 
& InternVL3-2B & 2B & 0.3807 & 0.3851 & 0.5821 & 0.4637 & 0.3077 \\
& InternVL3-8B & 8B & 0.5772 & 0.5696 & 0.7661 & 0.7371 & 0.4609 \\
& InternVL3-78B & 78B & 0.6443 & 0.6486 & 0.8444 & 0.7874 & 0.4630 \\
& Qwen3-VL-4B & 4B & 0.6261 & 0.6407 & 0.7503 & 0.7588 & 0.4909 \\
& Qwen3-VL-8B & 8B & 0.6545 & 0.6664 & 0.8577 & 0.7936 & 0.4792 \\
& Qwen3-VL-32B & 32B & 0.6751 & 0.6645 & 0.8310 & 0.8690 & 0.4852 \\
\hline
\multirow{4}{*}{Specialized} 
& DianJin-OCR-R1 & 7B & 0.4979 & 0.5038 & 0.7126 & 0.5992 & 0.2982 \\
& DeepSeek-OCR-small & 3B & 0.3959 & 0.3688 & 0.5081 & 0.5517 & 0.3637 \\
& olmOCR-7B & 7B & 0.6228 & 0.5736 & 0.8030 & 0.7854 & 0.4849 \\
& PaddleOCR-VL & 0.9B & 0.6344 & 0.6133 & 0.8125 & 0.8067 & 0.4746 \\
\hline
\multirow{6}{*}{\textbf{Ours (Fine-tuned)}} 
& InternVL3-2B (SFT) & 2B & 0.6496 & 0.7077 & 0.8514 & 0.7314 & 0.4509 \\
& InternVL3-2B (SFT+GRPO) & 2B & 0.4466 & 0.6279 & 0.4696 & 0.3954 & 0.2995 \\
& Qwen3-VL-4B (SFT) & 4B & 0.7723 & 0.8003 & 0.8964 & 0.8184 & 0.6478 \\
& Qwen3-VL-4B (SFT+GRPO) & 4B & 0.7788 & 0.8226 & 0.9298 & 0.8417 & 0.6215 \\
& \textbf{Qwen3-VL-8B (SFT)} & 8B & 0.7736 & 0.8155 & 0.9180 & 0.8273 & \textbf{0.6462} \\
& \textbf{Qwen3-VL-8B (SFT+GRPO)} & 8B & \textbf{0.7950} & \textbf{0.8488} & \textbf{0.9416} & 0.8547 & 0.6373 \\
\hline
\end{tabular}
}
\caption{\textbf{Main evaluation results (F1-score) on \datasetname.} Our fine-tuned models significantly outperform general baselines. While GRPO enhances overall performance for Qwen3-VL models (4B/8B) by boosting perception and reasoning, it poses stability challenges for the smaller InternVL3-2B.}
\vspace{-3mm}
\label{tab:main_results}
\end{table*}

\section{Experiments}
\label{sec:experiments}

\subsection{Experimental Setup}

\paragraph{Data Splitting.}
To ensure a rigorous evaluation that reflects the diversity of real-world scenarios, we partition the \datasetname\ dataset using stratified sampling based on receipt types. We reserve 2,000 images as the held-out test set to strictly maintain the distributional consistency with the full dataset. The remaining images are utilized for training.

\paragraph{Baselines.} We compare our fine-tuned models against a comprehensive set of state-of-the-art models, categorized into three groups: (1) \textbf{General Proprietary MLLMs}, represented by GPT-5 and Gemini-3-Pro; (2) \textbf{Open-source General MLLMs}, including the Qwen3-VL\citep{bai2025qwen3} and InternVL3\citep{zhu2025internvl3} series; and (3) \textbf{Specialized Document Models}, such as DianJin-OCR-R1, DeepSeek-OCR, PaddleOCR-VL, and olmOCR-7B~\citep{chen2025dianjin, wei2025deepseek, cui2025paddleocr, poznanski2025olmocr}.



\paragraph{Implementation Details.}
For our fine-tuned baselines, we utilize \textbf{Qwen3-VL-4/8B} and \textbf{InternVL3-2B} as backbones due to their efficiency. For detailed training configurations, hyperparameter settings and infrastructure specifications in Appendix \ref{app:hyperparams}.

\subsection{Main Results}

As shown in Table \ref{tab:main_results}, the results demonstrate that domain-specific alignment is a more decisive factor than raw parameter scale. Our fine-tuned \textbf{Qwen3-VL-8B} achieves an overall F1-score of \textbf{0.7950}, significantly outperforming proprietary state-of-the-art models like Gemini-3-Pro (0.7373) and GPT-5 (0.7076). This trend extends to data efficiency, where the compact \textbf{InternVL3-2B (SFT)} (0.6496) rivals the one-shot performance of the massive InternVL3.5-241B (0.6742).
While Metric-Aware GRPO successfully boosts the overall performance of Qwen3-VL models (e.g., 8B improves from 0.7736 to 0.7950), it poses stability challenges for the smaller InternVL3-2B. In the training logs, we observe \textit{reward collapse} and \textit{policy drift} evidenced by an initial reward increase that sharply declines around steps 220--230, accompanied by a significant spike in KL divergence. This suggests a capacity threshold for effective RL alignment, as 2B-scale models struggle to balance complex structural constraints against fundamental linguistic coherence. Conversely, specialized document models like DeepSeek-OCR and olmOCR exhibit significant performance drops in reasoning and structure generation, revealing that strong perceptual capabilities alone are insufficient for complex logic extraction. Furthermore, parsing nested structures remains the bottleneck across all baselines. While GPT-5 scores only 0.4893 on this metric, our SFT approach substantially improves this to \textbf{0.6478} (Qwen3-VL-4B), validating the effectiveness of our pipeline in handling heterogeneous layouts.

To ensure these findings are not skewed by the dominance of English or specific receipt types, we further evaluated our models on a curated category-balanced test set and a non-English subset. As detailed in Appendix \ref{app:robustness}, the relative performance rankings remain strictly consistent, demonstrating the cross-lingual and cross-category robustness of our framework.




\begin{figure}[t]
    \centering
    \setlength{\abovecaptionskip}{0.2cm}
    \setlength{\belowcaptionskip}{-0.3cm}
    \includegraphics[width=0.48\textwidth]{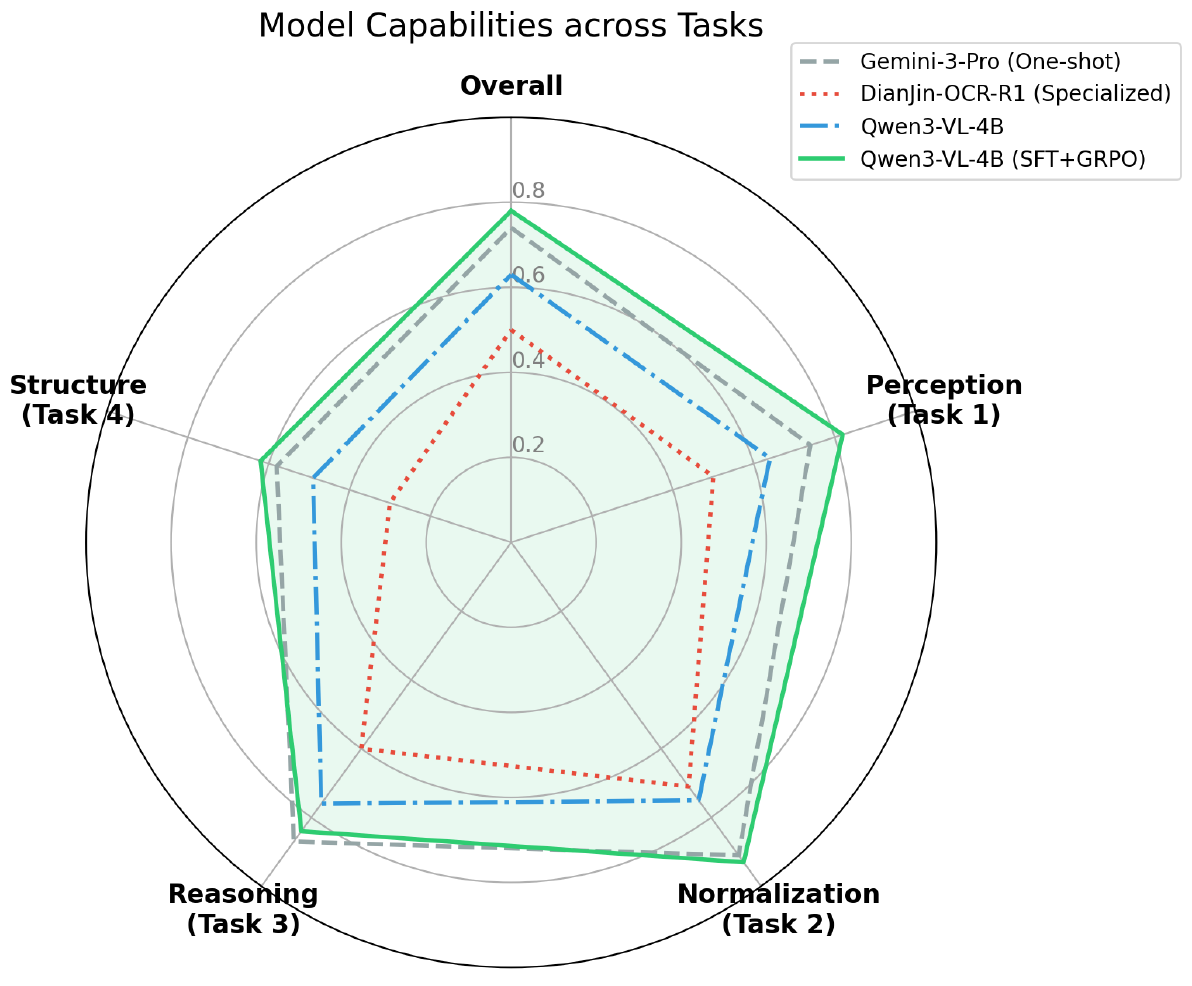}
    \caption{\textbf{Holistic Evaluation.} The chart compares model capabilities across the four sub-tasks. While proprietary models (gray) are balanced, our fine-tuned baseline (green) excels in domain-specific structure parsing.}
    \vspace{-3mm}
    \label{fig:radar}
\end{figure}

\subsection{Analysis}
\label{sec:error_analysis}

\begin{figure}[t]
    \setlength{\abovecaptionskip}{0.2cm}
    \setlength{\belowcaptionskip}{-0.1cm}
    \centering
    \begin{subfigure}{1.0\linewidth}
        \centering
        \includegraphics[width=\linewidth]{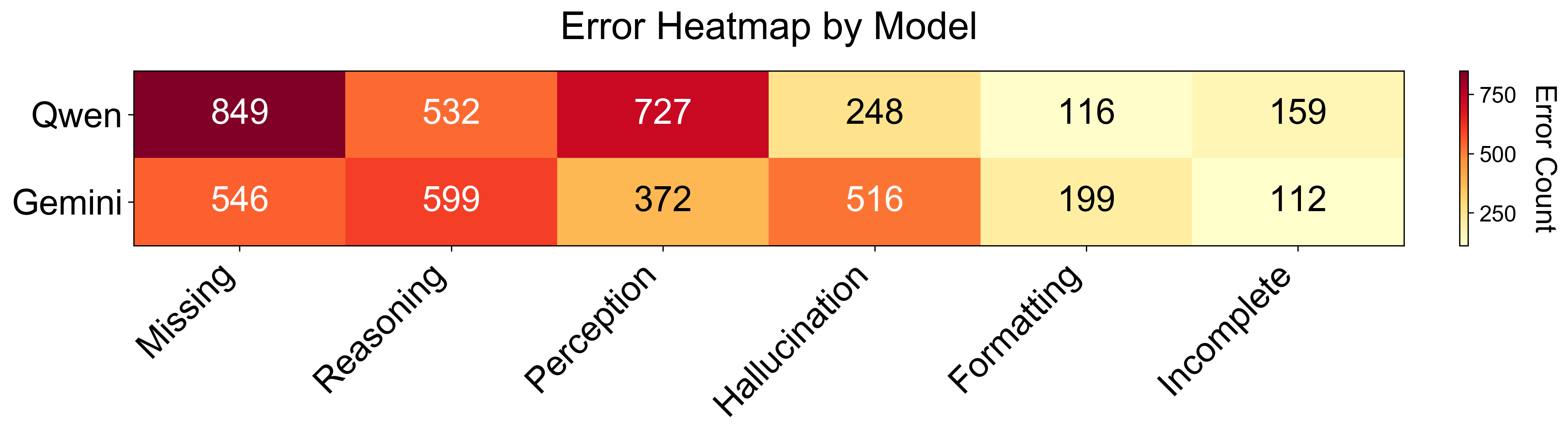}
        \caption{\textbf{Error Type Distribution.} Qwen3-VL (Ours) shows a "conservative" pattern with high \textit{Missing} rates (849), whereas Gemini-3-Pro is "aggressive" with high \textit{Hallucination} (516) and \textit{Reasoning} errors.}
        \label{fig:error_heatmap}
    \end{subfigure}

    \par\bigskip 
    
    \begin{subfigure}{1.0\linewidth}
        \centering
        \includegraphics[width=\linewidth]{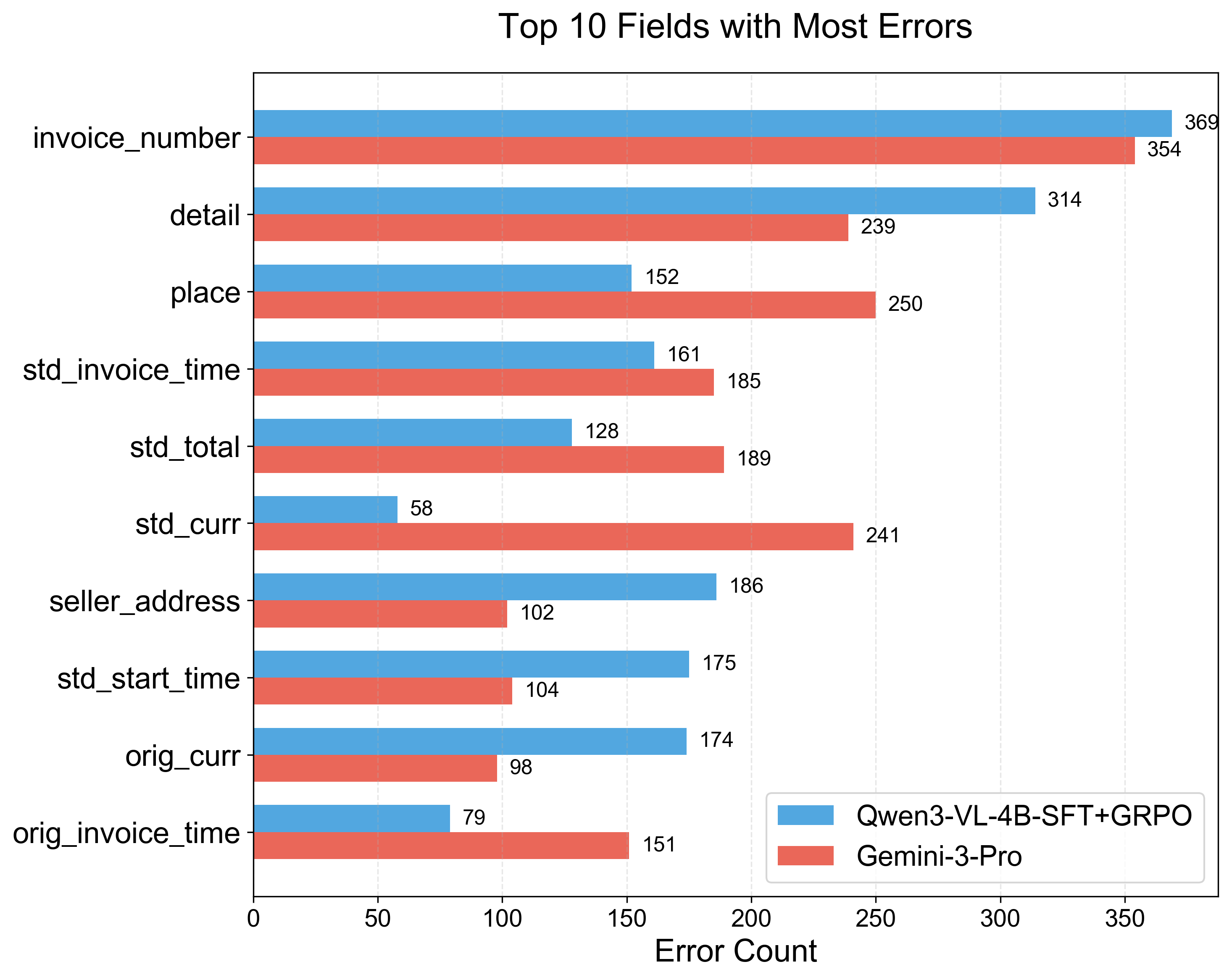}
        \caption{\textbf{Top 10 Error-Prone Fields.} \texttt{invoice\_number} and \texttt{detail} are the hardest fields. Note the significant gap in \texttt{place}, indicating Gemini's tendency to over-interpret location context.}
        \label{fig:error_barchart}
    \end{subfigure}
    
    \caption{\textbf{Fine-grained Error Analysis on \datasetname.} We compare the error patterns of our fine-tuned Qwen3-VL-4B against Gemini-3-Pro. (a) illustrates the divergent behavioral profiles, while (b) highlights the specific fields that pose the greatest challenges.}
    \vspace{-3mm}
    \label{fig:error_analysis_combined}
\end{figure}

To understand the capability boundaries of current MLLMs, we conducted a fine-grained error analysis comparing our fine-tuned \textbf{Qwen3-VL-4B} against the proprietary \textbf{Gemini-3-Pro}. As illustrated in Figure \ref{fig:error_analysis_combined}, we categorize the failure modes into three distinct patterns.

\paragraph{Perception Bottlenecks: Visual Ambiguity vs. Hallucination.}


As illustrated in Figure \ref{fig:error_analysis_combined}, the perception task reveals a fundamental behavioral divergence between the two models. Qwen3-VL suffers predominantly from \textit{Missing} errors (849 cases) and \textit{Perception} errors (727 cases). It struggles with fine-grained OCR in dense layouts, often failing to detect fields like \texttt{orig\_curr} or misidentifying confusing digits (e.g.,"1"/"7") in \texttt{invoice\_number}—which ranks as the most error-prone field for both models (Figure \ref{fig:error_analysis_combined}b). Conversely, Gemini exhibits a high tendency for \textit{Hallucination} (Gemini: 516 cases vs. Qwen: 248). When a unique ID is visually absent, Gemini often fabricates a plausible string for \texttt{invoice\_number} to satisfy the schema, rather than outputting an empty string. This highlights the challenge of grounding generation strictly in visual evidence.


\paragraph{Reasoning Gaps: Contextual Inference.}



Beyond perception, \textbf{Reasoning and Normalization} tasks account for the majority of Gemini's failures (599 Reasoning errors) exposing critical deficiencies in utilizing global context. The Field Error Ranking (Figure \ref{fig:error_analysis_combined}b) highlights a massive performance gap in the \texttt{place} field (Gemini: 250 vs. Qwen: 152). Gemini frequently hallucinates specific cities based on currency cues (e.g., inferring "London" from "£"), whereas Qwen tends to remain conservative when the address is ambiguous. Similarly, in \texttt{std\_curr}, Gemini produces significantly more errors (241 vs. 58). Models often default to USD when the symbol "\$" is ambiguous, failing to cross-reference the \texttt{seller\_address} to correctly infer CAD or AUD. Ambiguous date formats (e.g., "02/03/24") lead to swapping Day/Month. Gemini's higher error count in \textit{Formatting} (199 cases) suggests it often ignores specific normalization instructions compared to the fine-tuned Qwen.

\paragraph{Consistency Trap in Structural Parsing.}
The \texttt{detail} field ranks as the second most difficult field in Figure \ref{fig:error_analysis_combined}b. A critical finding in the \textbf{Structure} task is the phenomenon of \textit{"Hallucination for Arithmetic Consistency."} 
Complex invoices imply constraints (e.g., $\sum \text{items} = \text{Total}$). Models, particularly Gemini, often tamper with visual data to satisfy these priors: \textbf{Value Tampering:} To force the sum of \texttt{detail} items to match the \texttt{std\_total}, models occasionally alter the price of a line item or hallucinate a non-existent "Tax" item. \textbf{Unwanted Calculation:} When the total amount is visually missing, models attempt to manually sum up line items to generate a \texttt{std\_total}, leading to calculation errors.


This underscores the value of our Metric-Aware GRPO arithmetic reward, which enforces logical consistency without compromising visual faithfulness.

\subsection{Ablation Studies}

\paragraph{Effect of Training Stages.}
We analyze the contribution of each stage using Qwen3-VL-4B. As shown in Table \ref{tab:ablation_training}, \textbf{SFT} establishes a critical foundation, yielding massive gains in \textit{Perception} (+16.0\%) and \textit{Normalization} (+14.6\%) over the one-shot baseline by teaching schema adherence.
The introduction of \textbf{Metric-Aware GRPO} further refines these capabilities. Interestingly, "GRPO Only" achieves the highest \textit{Reasoning} score (0.8560), indicating RL's potency in optimizing logic, yet it lags in visual grounding. Consequently, the combined \textbf{SFT + GRPO} strategy achieves the optimal balance, delivering state-of-the-art results in \textit{Perception} (\textbf{0.8226}) and \textit{Normalization} (\textbf{0.9298}) while maintaining strong reasoning gains. Crucially, this RL alignment explicitly suppresses hallucinations. Quantitative analysis confirms that Metric-Aware GRPO reduces False Positives (FPs) by up to 68.9\% in complex fields while significantly boosting overall precision (see Appendix \ref{app:hallucination}).

\begin{table}[h]
\centering
\small
\begin{tabular}{l|ccc}
\hline
\textbf{Stage} & \textbf{Perception} & \textbf{Normalization} & \textbf{Reasoning} \\
\hline
One-shot & 0.6407 & 0.7503 & 0.7588 \\
SFT Only & 0.8003 & 0.8964 & 0.8184 \\
GRPO Only & 0.7758 & 0.8782 & \textbf{0.8560} \\
SFT + GRPO & \textbf{0.8226} & \textbf{0.9298} & 0.8417 \\
\hline
\end{tabular}
\caption{Ablation of training stages. SFT enables robust visual grounding and formatting, while GRPO is essential for maximizing reasoning capabilities.}
\vspace{-3mm}
\label{tab:ablation_training}
\end{table}

\section{Conclusion}

We introduce \textbf{\datasetname}, a benchmark designed to propel Visual Information Extraction (VIE) from literal extraction toward cognitive reasoning. Moving beyond retail-centric datasets, \datasetname\ comprises 10k diverse overseas financial documents with a hierarchical taxonomy covering Perception, Normalization, Reasoning, and Structure. 
To tackle these challenges, we propose a two-stage training framework combining SFT with \textbf{Metric-Aware GRPO}. Experiments demonstrate that while SFT establishes a solid foundation, our RL alignment significantly mitigates hallucinations and improves arithmetic consistency. 
However, the persistent performance gap in structural parsing highlights that handling nested layouts remains an open research problem. We hope \datasetname\ serves as a rigorous testbed for next-generation multimodal agents, fostering advancements in autonomous financial auditing.

\section*{Limitations}

While \datasetname\ represents a significant step forward, it has certain limitations. 

First, although the dataset covers multiple languages, it is predominantly English-centric ($97.9\%$), reflecting the data availability in open web sources. Future work should focus on scaling low-resource languages to improve multilingual robustness.

Second, to strictly protect privacy, all PII (Personally Identifiable Information) was masked. While necessary, this may slightly alter the visual distribution compared to raw private financial data found in internal corporate streams.

Third, we did not perform systematic visual data augmentation (e.g., rotation, gaussian blur, or noise injection) during the evaluation. While our dataset contains natural visual variations from real-world collection, we have not explicitly stress-tested the models' robustness against severe visual degradations or adversarial attacks.

Finally, our proposed GRPO training method, while effective, incurs a higher computational cost compared to standard SFT. Developing more data-efficient alignment strategies for MLLMs remains a valuable direction for future exploration.

\section*{Ethics Statement}
Given that our data originates from real-world transactions, we enforced a strict de-identification policy where sensitive PII (e.g., personal names) was detected and masked with \textbf{irreversible black boxes} during annotation, ensuring effective anonymity by rendering private information visually and digitally inaccessible.

\section*{Acknowledgments}
This work was supported in part by the Ningbo Youth Science and Technology Innovation Leading Talent Program (No. 2025QL059), CCF-1688 Yuanbao Collaborative Fund (No. CCF-Alibaba 2025004), the "Pioneer and Leading Goose" R\&D Program of Zhejiang (No. 2025C02037), the Zhejiang Provincial Philosophy and Social Sciences Planning Project (No. 22QNYC04ZD), the National Social Science Fund of China (No. 24BGL071), and the Fundamental Research Funds for the Central Universities.

We gratefully acknowledge Ziman Li for her assistance in designing the annotation schema and guidelines; Xiaoqing Liu, Yongbo Wang, and Lufei Xu for their help with receipt image collection; Enci Zhang, Xiang Li, Wuyou Mao, Yingtian Hu, and Shujian Zhu for their contributions to annotated data validation; and Qi Yang and Yuan Liu, together with the above validation team, for error analysis during model iterations.

\bibliography{custom}

\clearpage

\appendix

\begin{table*}[t]
\setlength{\abovecaptionskip}{0.2cm}
\setlength{\belowcaptionskip}{-0.3cm}
\centering
\small
\renewcommand{\arraystretch}{1.2} 
\resizebox{\textwidth}{!}{%
\begin{tabular}{l|l|l|c}
\hline
\textbf{Field Name} & \textbf{Definition} & \textbf{Sub-task Category} & \textbf{Metric} \\ \hline
\texttt{orig\_start\_time} & Start time of service as it appears visually (raw text) & Basic Perception & Semantic \\
\texttt{orig\_end\_time} & End time of service as it appears visually (raw text) & Basic Perception & Semantic \\
\texttt{orig\_invoice\_time} & Issuance time as it appears visually (raw text) & Basic Perception & Semantic \\
\texttt{orig\_total} & Total amount as it appears visually (raw text) & Basic Perception & Numeric \\
\texttt{orig\_curr} & Currency clues like symbol or city \& country as it appears visually (e.g., \$) & Basic Perception & Structured List \\ 
\texttt{invoice\_number} & Unique identifier of the receipt/invoice & Basic Perception & Exact Match \\
\texttt{tax\_number} & Tax identification number of the merchant & Basic Perception & Exact Match \\
\texttt{seller\_name} & Name of the merchant or service provider & Basic Perception & Semantic \\ \hline
\texttt{std\_start\_time} & Start time normalized to YYYY-MM-DD format & Formatting \& Normalization & Exact Match \\
\texttt{std\_end\_time} & End time normalized to YYYY-MM-DD format & Formatting \& Normalization & Exact Match \\
\texttt{std\_invoice\_time} & Issuance time normalized to YYYY-MM-DD format & Formatting \& Normalization & Exact Match \\
\texttt{std\_total} & Total amount normalized to decimal format (e.g., 1,000.00) & Formatting \& Normalization & Numeric \\ \hline
\texttt{type} & Classification of expense (e.g., Hotel, Train, Taxi) & Semantic Reasoning & Exact Match \\
\texttt{place} & Location where the expense occurred & Semantic Reasoning & Semantic \\
\texttt{departure} & Origin city (for transport tickets) & Semantic Reasoning & Semantic \\
\texttt{arrival} & Destination city (for transport tickets) & Semantic Reasoning & Semantic \\
\texttt{std\_curr} & Standardized ISO currency code inferred from context (e.g., USD) & Semantic Reasoning & Exact Match \\
\texttt{seller\_address} & City that the merchant locates & Semantic Reasoning & Semantic \\ \hline
\texttt{detail} & Structured list of line items (content, amount, tax status) & Structural Parsing & Structured List \\ \hline
\end{tabular}%
}
\caption{Basic definitions, categories and metrics of the 19 annotation fields in our dataset. These fields are categorized into four sub-tasks based on the required cognitive capability (see Section~\ref{sec:schema}). The evaluation metrics are defined in Section~\ref{sec:eval_protocol}.}
\label{tab:fields_definition}
\end{table*}

\begin{table}[h]
\setlength{\abovecaptionskip}{0.1cm}
\setlength{\belowcaptionskip}{-0.4cm}
\centering
\footnotesize 
\renewcommand{\arraystretch}{0.95} 
\small
\begin{tabular}{l|r|r}
\hline
\textbf{Language} & \textbf{Count} & \textbf{Pct.} \\ \hline
\textbf{English} & 10,443 & 98.00\% \\ \hline
French & 60 & 0.56\% \\
Spanish & 51 & 0.49\% \\
German & 31 & 0.29\% \\
Indonesian & 31 & 0.29\% \\
Portuguese & 18 & 0.18\% \\
Romanian & 10 & 0.09\% \\
Others & 11 & 0.10\% \\ \hline
\textbf{Non-English Total} & \textbf{213} & \textbf{2.00\%} \\ \hline
\textbf{Total} & \textbf{10,656} & \textbf{100.0\%} \\ \hline
\end{tabular}%
\caption{Language distribution of the dataset. "Others" includes Italian, Korean, and Japanese.}
\vspace{-3mm}
\label{tab:lang_distribution}
\end{table}

\section{Dataset Details \& Field Specifications}
\label{annoSchema}

Our dataset defines 19 fields designed to provide a comprehensive evidence chain for financial auditing. The selection of these fields is grounded in the \textit{Generally Accepted Accounting Principles (GAAP)} \citep{fasb_gaap} and international tax regulations (e.g., EU VAT Directive \citep{eu_vat_directive}), ensuring the benchmark's utility for real-world financial auditing. The annotation rules for each field are detailed below.

\paragraph{1. Entity Verification}
This dimension focuses on identifying the stakeholders involved in the transaction to establish legitimacy.

\begin{itemize}
    \item \textbf{seller\_name}: The name of the merchant or service provider. As these refer to public business entities, they are not considered PII. The annotation must be faithful to the visual information on the receipt (e.g., logos, headers). 
    \item \textbf{seller\_address}: The city that the merchant located, formatted as ``Country-City'' (e.g., \textit{UK-London}). Inferring addresses via external search engines is strictly prohibited to ensure the dataset reflects only the information contained in the image.
    \item \textbf{invoice\_number}: The unique identifier of the receipt or invoice. Common labels include ``Invoice No.'', ``Receipt No.'', ``Confirmation No.'' or ``Ticket No.''. If multiple numbers exist (e.g., Order No. and Invoice No.), the Invoice Number takes precedence as the primary financial identifier.
    \item \textbf{tax\_number}: The tax identification number of the merchant (e.g., VAT ID, GST No., TIN).
\end{itemize}

\paragraph{2. Financial Integrity}
This dimension captures critical financial data to verify calculations and amounts.

\begin{itemize}
    \item \textbf{orig\_total}: The total amount of the transaction as it appears visually in the raw text. This field captures the exact string from the document, including original separators (e.g., \texttt{1.000,00}), without any normalization.
    \item \textbf{std\_total}: The normalized total amount for computational verification. The value is standardized to a decimal format with two decimal places (e.g., \texttt{1,000.00}). Thousands separators are unified to commas. Logic rules dictate that this should be the final amount payable, inclusive of taxes and tips.
    \item \textbf{orig\_curr}: Visual evidence of the currency. This includes symbols (e.g., \$, \texteuro), text abbreviations (e.g., USD, RMB), or geographic clues (e.g., ``Toronto'' implying Canadian Dollars) explicitly found on the image.
    \item \textbf{std\_curr}: The standardized 3-letter ISO currency code (e.g., USD, EUR, GBP, CNY). This is inferred from the \texttt{orig\_curr} evidence.
    \item \textbf{detail}: A structured list containing line items to verify the breakdown of the total amount. This is a complex field where each item is a JSON object containing three sub-components:
    \begin{itemize}
        \item \texttt{content}: The description of the product or service.
        \item \texttt{amount}: The numerical value of the specific item.
        \item \texttt{ifTax}: A boolean flag (\texttt{True}/\texttt{False}) indicating whether the item represents a tax charge (e.g., VAT, GST).
    \end{itemize}
    Annotators ensure that the summation of these line items logically aligns with the \texttt{std\_total}.
\end{itemize}

\paragraph{3. Spatio-Temporal Validation}
This dimension validates when and where the expense occurred to ensure the context matches the business trip or transaction claim.

\begin{itemize}
    \item \textbf{place}: The location where the expense occurred, formatted as ``Country-City'' (e.g., \textit{UK-London}). If the document only specifies a city, the country is added; if only the country is visible, the city is left blank.
    \item \textbf{departure}: The origin city for transportation tickets. This applies to cross-city travel (plane, train, bus). If a trip involves multiple segments (e.g., A-B-A), only the initial departure point is recorded.
    \item \textbf{arrival}: The destination city for transportation tickets. Similar to \texttt{departure}, this captures the endpoint of the travel service.
    \item \textbf{orig\_start\_time}: The raw text indicating the start of the service or event. It preserves the original date format found on the image (e.g., ``15-July-24'').
    \item \textbf{std\_start\_time}: The normalized start date converted to the ISO \texttt{YYYY-MM-DD} format (e.g., \texttt{2024-07-15}). This facilitates temporal reasoning. Logic rules handle ambiguous formats (e.g., 07/06/24) by cross-referencing the country's date convention.
    \item \textbf{orig\_end\_time}: The raw text indicating the end of the service (e.g., hotel check-out, flight arrival). If the transaction occurs on a single day, this field should be left empty.
    \item \textbf{std\_end\_time}: The normalized end date converted to \texttt{YYYY-MM-DD}. 
    \item \textbf{orig\_invoice\_time}: The raw text indicating when the invoice/receipt was issued. For on-the-spot receipts (e.g., retail receipts), this is identical to the transaction time; for post-paid invoices, it may differ from the service period.
    \item \textbf{std\_invoice\_time}: The normalized issuance date converted to \texttt{YYYY-MM-DD}. 
\end{itemize}

\paragraph{4. Expense Classification}
This dimension categorizes the nature of the transaction for accounting and reimbursement purposes.

\begin{itemize}
    \item \textbf{type}: A classification label selected from a standardized list: \textit{plane, train, ship, bus, taxi, metro, hotel}, or \textit{other}. Annotators determine this based on explicit keywords (e.g., ``Flight'' $\rightarrow$ \textit{plane}) or implicit logic (e.g., ``Double Room'' $\rightarrow$ \textit{hotel}).
\end{itemize}

Table~\ref{tab:fields_definition} shows the basic definitions, sub-task categories, and metrics of these 19 fields. Table~\ref{tab:lang_distribution} shows the language distribution of the dataset.

\section{Implementation Details}
\label{app:hyperparams}
\subsection{Implementation Details and Hyperparameters}
We utilized the LLaMA-Factory framework \citep{zheng2024llamafactory} to fine-tune the \textbf{Qwen3-VL} series and \textbf{InternVL-3} models. The training configurations were set as follows: 
In the \textbf{SFT stage}, models are trained for 2 epochs with a global batch size of 16 (achieved via gradient accumulation steps of 8 on single-device batches), a learning rate of $1e-5$ with a cosine decay scheduler, and BF16 mixed-precision. Notably, we set the maximum context length to \textbf{5,120 tokens} to accommodate receipts with long lists of items (the \texttt{detail} field), ensuring no information truncation during training.
In the \textbf{GRPO stage}, we employ the reward function defined in Section \ref{sec:method}, setting the KL coefficient to 0.01 and collecting 16 samples per prompt for policy updates. All experiments are conducted on 4$\times$NVIDIA A800 GPUs.

\section{Evaluation Details}
\subsection{Prompts for Instruction Tuning and Inference}
\label{app:prompts}

To ensure the model adheres to the strict output schema required by \datasetname, we designed a comprehensive system prompt. Table \ref{tab:full_prompt} illustrates the exact prompt used during both the Supervised Fine-Tuning (SFT) and inference stages. The prompt consists of three components: (1) a role definition and format constraint, (2) detailed extraction rules for each field, and (3) a one-shot demonstration to guide the JSON structure.

\begin{table*}[h!]
\centering
\small
\renewcommand{\arraystretch}{1.1}
\begin{tabular}{|p{0.96\textwidth}|}
\hline
\textbf{\textsf{System Instruction}} \\
\hline
You are an AI assistant specialized in extracting structured information from images of overseas travel and expense invoices. Your task is to analyze each invoice image and convert the relevant information into a structured JSON object.
Follow the field definitions and annotation rules below precisely. The output must be a valid JSON object containing all required key-value pairs, using exact formats as described.

\vspace{0.5em}
\textbf{\textsf{OUTPUT FORMAT}} \\
Your output must be a single \textbf{valid JSON object} for each invoice. Do not include any explanatory text or formatting like markdown.
Below are the output rules for the key-value pairs in the json objects. Specifically, if a field is missing or cannot be identified, return it as the empty string \texttt{""}.

\vspace{0.5em}
\textbf{\textsf{EXTRACTION RULES}} \\
\begin{itemize}[leftmargin=1.5em, topsep=0pt, itemsep=0pt]
    \item \texttt{type}: output a string, infer from keywords/context and choose within ["plane", "train", "ship", "bus", "taxi", "metro", "hotel", "other"].
    \item \texttt{orig\_start\_time}: output a string, start time of the event. Keep original format.
    \item \texttt{orig\_end\_time}: output a string, end time of the event. Keep original format. Return empty if within 1 day.
    \item \texttt{orig\_invoice\_time}: output a string, date of issuance. Evidence needed. Keep original format.
    \item \texttt{std\_start\_time}: output a string, "YYYY-MM-DD" format of \texttt{orig\_start\_time}.
    \item \texttt{std\_end\_time}: output a string, "YYYY-MM-DD" format of \texttt{orig\_end\_time}.
    \item \texttt{std\_invoice\_time}: output a string, "YYYY-MM-DD" format of \texttt{orig\_invoice\_time}.
    \item \texttt{place}: output a string, "Country-City" format. Only valid if clearly stated.
    \item \texttt{departure}: output a string, "Country-City" format. Only for intercity travel.
    \item \texttt{arrival}: output a string, "Country-City" format. Only for intercity travel.
    \item \texttt{orig\_curr}: output a list of evidence strings (e.g., "\$", "Toronto"). Keep original content.
    \item \texttt{std\_curr}: output a string, 3-letter standard ISO code (e.g. USD, EUR, CAD).
    \item \texttt{orig\_total}: output a string, total amount. Keep original content. No symbol needed.
    \item \texttt{std\_total}: output a string, standard format with comma separator, 2 decimal places (e.g. "1,200.00").
    \item \texttt{detail}: output a list of dicts: \texttt{[ \{"content": "...", "amount": "...", "ifTax": True/False\} ]}. Exclude subtotals.
    \item \texttt{seller\_name}: output a list of seller names. Valid only when explicitly present.
    \item \texttt{seller\_address}: output a list of seller cities in ["Country-City"] format.
    \item \texttt{invoice\_number}: output a string. Strip prefixes like "No.". Prefer official IDs.
    \item \texttt{tax\_number}: output a string. Strip prefixes like "Tax ID".
\end{itemize}

\vspace{0.5em}
\textbf{\textsf{FEW-SHOT EXAMPLE}} \\
\texttt{\{} \\
\texttt{\ \ "type": "train",} \\
\texttt{\ \ "std\_start\_time": "2024-07-06", "orig\_start\_time": "06 Jul 2024",} \\
\texttt{\ \ "std\_end\_time": "", "orig\_end\_time": "",} \\
\texttt{\ \ "std\_invoice\_time": "2024-06-03", "orig\_invoice\_time": "03 Jun 2024",} \\
\texttt{\ \ "place": "Australia-Sydney",} \\
\texttt{\ \ "departure": "Australia-Sydney", "arrival": "Australia-Canberra",} \\
\texttt{\ \ "std\_curr": "AUD", "orig\_curr": ["\$", "Sydney"],} \\
\texttt{\ \ "std\_total": "50.58", "orig\_total": "50.58",} \\
\texttt{\ \ "detail": [} \\
\texttt{\ \ \ \ \{ "content": "Trip Fare", "amount": "45.00", "ifTax": False \},} \\
\texttt{\ \ \ \ \{ "content": "Tax fee", "amount": "5.58", "ifTax": True \}} \\
\texttt{\ \ ],} \\
\texttt{\ \ "seller\_name": ["NSW TrainLink"],} \\
\texttt{\ \ "seller\_address": ["Australia-Sydney"],} \\
\texttt{\ \ "invoice\_number": "0306202450122",} \\
\texttt{\ \ "tax\_number": "50 325 560 455"} \\
\texttt{\}} \\

\vspace{0.5em}
\textbf{\textsf{USER INPUT}} \\
Extract now (JSON only, no explanation): \\
\hline
\end{tabular}
\caption{The full system prompt used for SFT and inference on \datasetname. The prompt enforces schema constraints, defines normalization rules, and provides a one-shot demonstration to guide the model's output format.}
\label{tab:full_prompt}
\end{table*}

\subsection{Prompt for LLM Semantic Judge}
\label{app:prompt_judge}

For fields requiring semantic reasoning (e.g., \texttt{place}, \texttt{seller\_name}), we employ a lightweight LLM as a judge when exact matching fails. Table \ref{tab:judge_prompt} details the instruction provided to the judge model to determine semantic equivalence.

\begin{table*}[h]
\centering
\small
\renewcommand{\arraystretch}{1.1}
\begin{tabular}{|p{0.96\textwidth}|}
\hline
\textbf{\textsf{System Instruction}} \\
You are an expert data quality analyst. Your task is to determine if the 'Predicted Value' is semantically equivalent to the 'Ground Truth Value' for a specific field extracted from a document. \\

\vspace{0.3em}
\textbf{\textsf{Context}} \\
- Field Name: \texttt{<field\_name>} \\

\vspace{0.3em}
\textbf{\textsf{Equivalence Criteria}} \\
Consider the values \textbf{equivalent} if they represent the same real-world entity or meaning, even with minor differences like:
\begin{itemize}[leftmargin=1.5em, topsep=2pt, itemsep=0pt]
    \item Abbreviations (e.g., "Co." vs. "Company").
    \item Common synonyms or alternative names.
    \item Minor typos or spelling errors that do not change the meaning.
    \item Formatting differences (e.g., "1,234.50" vs. "1234.50").
    \item Presence or absence of trivial words (e.g., "The Grand Hotel" vs. "Grand Hotel").
\end{itemize}

Consider the values \textbf{NOT equivalent} if:
\begin{itemize}[leftmargin=1.5em, topsep=2pt, itemsep=0pt]
    \item They refer to different entities (e.g., "Pepsi" vs. "Coca-Cola").
    \item The core information is different (e.g., a different address or name).
    \item The prediction contains significant missing or extra information that changes the meaning.
\end{itemize}

\vspace{0.3em}
\textbf{\textsf{Task}} \\
Based on the criteria above, evaluate the following pair:
\begin{itemize}[leftmargin=1.5em, topsep=2pt, itemsep=0pt]
    \item Ground Truth Value: "\texttt{<ground\_truth>}"
    \item Predicted Value: "\texttt{<prediction>}"
\end{itemize}

\vspace{0.3em}
\textbf{\textsf{Output}} \\
Respond ONLY with a valid JSON object containing two keys:
\begin{enumerate}[leftmargin=2em, topsep=2pt, itemsep=0pt]
    \item \texttt{"is\_equivalent"}: A boolean value (true or false).
    \item \texttt{"reasoning"}: A brief explanation for your decision.
\end{enumerate} \\
\hline
\end{tabular}
\caption{The full prompt used for the LLM-based Semantic Judge. This prompt is triggered only when the Levenshtein similarity between the prediction and ground truth falls below the exact match threshold.}
\label{tab:judge_prompt}
\end{table*}

\section{Additional Results}
\subsection{Robustness across Languages and Categories}
\label{app:robustness}

To address potential evaluation biases arising from data distribution, we conducted robustness checks on two specific subsets: a curated category-balanced test set and a non-English subset.

\paragraph{Category-Balanced Evaluation.} We constructed a balanced test set comprising 1,387 samples by down-sampling dominant categories (e.g., Purchase, Dining) to match the frequency of minority classes. As shown in Table \ref{tab:balanced_test}, while the absolute F1 scores shifted slightly due to the altered distribution, the relative ranking of the models remained highly consistent, with our SFT+GRPO framework maintaining its superior performance.

\begin{table}[h]
\centering
\small
\resizebox{\columnwidth}{!}{%
\begin{tabular}{l|c|cccc}
\hline
\textbf{Model} & \textbf{Overall} & \textbf{Perc.} & \textbf{Norm.} & \textbf{Reason.} & \textbf{Struct.} \\ \hline
Gemini-3-Pro & 0.6863 & 0.6817 & 0.9022 & 0.8019 & 0.5481 \\
Qwen3-VL-8B & 0.6544 & 0.6690 & 0.8554 & 0.7975 & 0.4770 \\
Qwen3-VL-8B (SFT) & 0.7639 & 0.8130 & 0.9138 & 0.8285 & \textbf{0.6366} \\
Qwen3-VL-8B (SFT+GRPO) & \textbf{0.7861} & \textbf{0.8478} & \textbf{0.9377} & \textbf{0.8548} & 0.6112 \\ \hline
\end{tabular}%
}
\caption{Evaluation results on the category-balanced test set.}
\label{tab:balanced_test}
\end{table}

\paragraph{Cross-Lingual Robustness.} We also evaluated the models on the 2\% non-English subset. As detailed in Table \ref{tab:non_english}, despite the data scarcity for these low-resource languages, our fine-tuned models exhibit significant improvements. Notably, the Qwen3-VL-8B (SFT+GRPO) achieves a leading F1 score of 0.7190, outperforming both GPT-5 (0.6441) and Gemini-3-Pro (0.6827). This demonstrates that our Metric-Aware GRPO method successfully enables the model to capture universal layout and structural patterns, effectively mitigating the impact of language barriers.

\begin{table}[h]
\centering
\small
\resizebox{\columnwidth}{!}{%
\begin{tabular}{l|c|cccc}
\hline
\textbf{Model} & \textbf{Overall} & \textbf{Perc.} & \textbf{Norm.} & \textbf{Reason.} & \textbf{Struct.} \\ \hline
GPT-5 & 0.6441 & 0.6968 & 0.8299 & 0.8112 & 0.3058 \\
Gemini-3-Pro & 0.6827 & 0.7123 & 0.8396 & 0.8092 & \textbf{0.4305} \\ \hline
Qwen3-VL-4B & 0.5906 & 0.6172 & 0.7593 & 0.7579 & 0.3143 \\
Qwen3-VL-4B (SFT) & 0.6574 & 0.7468 & 0.7212 & 0.7781 & 0.4081 \\
Qwen3-VL-4B (+GRPO) & 0.6832 & 0.7532 & 0.7681 & 0.8432 & 0.3429 \\ \hline
Qwen3-VL-8B & 0.6193 & 0.6504 & 0.7733 & 0.8011 & 0.2954 \\
Qwen3-VL-8B (SFT) & 0.6745 & \textbf{0.7920} & 0.7491 & 0.8011 & 0.3531 \\
Qwen3-VL-8B (+GRPO) & \textbf{0.7190} & 0.7912 & \textbf{0.8620} & \textbf{0.8441} & 0.3810 \\ \hline
\end{tabular}%
}
\caption{Evaluation results on the non-English subset.}
\label{tab:non_english}
\end{table}

\subsection{Quantitative Proof of Hallucination Suppression}
\label{app:hallucination}

Our Metric-Aware GRPO explicitly penalizes hallucinations through negative rewards for False Positives (FP). To quantitatively demonstrate this, we compared the Precision scores and the absolute FP counts before and after RL alignment.

As shown in Table \ref{tab:precision_improvement}, both Qwen3-VL-4B and 8B models exhibit significant improvements in Precision after GRPO training (e.g., the 8B model's overall Precision increased from 0.8319 to 0.8794). Furthermore, Table \ref{tab:fp_reduction} illustrates a consistent reduction in the absolute number of False Positives across various fields. Notably, hallucinated predictions for the complex \texttt{detail} field dropped by 20.0\%, and errors in \texttt{std\_invoice\_time} decreased sharply by 68.8\%. These quantitative results confirm that the performance gains of our framework are heavily driven by substantial hallucination suppression.

\begin{table}[h]
\centering
\small
\resizebox{\columnwidth}{!}{%
\begin{tabular}{l|c|cccc}
\hline
\textbf{Model (Precision)} & \textbf{Overall} & \textbf{Perc.} & \textbf{Norm.} & \textbf{Reason.} & \textbf{Struct.} \\ \hline
Qwen3-VL-4B (SFT) & 0.8134 & 0.8527 & 0.9329 & 0.9271 & 0.6756 \\
Qwen3-VL-4B (+GRPO) & 0.8693 & 0.9022 & 0.9662 & 0.9664 & 0.7479 \\ \hline
Qwen3-VL-8B (SFT) & 0.8319 & 0.8590 & 0.9349 & 0.9502 & 0.7351 \\
Qwen3-VL-8B (+GRPO) & \textbf{0.8794} & \textbf{0.9099} & \textbf{0.9690} & \textbf{0.9703} & \textbf{0.7664} \\ \hline
\end{tabular}%
}
\caption{Precision improvement after GRPO alignment.}
\label{tab:precision_improvement}
\end{table}

\begin{table}[h]
\centering
\small
\resizebox{\columnwidth}{!}{%
\begin{tabular}{l|cc|cc}
\hline
\textbf{Field Name} & \textbf{SFT (FP)} & \textbf{+GRPO (FP)} & \textbf{Change} & \textbf{Change (\%)} \\ \hline
\texttt{std\_invoice\_time} & 410 & 128 & -282 & -68.8\% \\
\texttt{invoice\_number} & 398 & 142 & -256 & -64.3\% \\
\texttt{arrival} & 53 & 27 & -26 & -49.1\% \\
\texttt{std\_curr} & 68 & 38 & -30 & -44.1\% \\
\texttt{seller\_address} & 147 & 116 & -31 & -21.1\% \\
\texttt{detail} & 2079 & 1664 & -415 & -20.0\% \\ \hline
\end{tabular}%
}
\caption{Reduction of False Positives (FP) for the Qwen3-VL-8B model across representative fields after Metric-Aware GRPO training.}
\label{tab:fp_reduction}
\end{table}

\subsection{Hyperparameter Sensitivity Analysis for Structural Similarity}
\label{app:sensitivity}

To ensure the structural parsing similarity score in Equation (1) robustly reflects true semantic understanding, the weights ($\alpha, \beta, \gamma, \delta$) were determined through a rigorous empirical validation process. Furthermore, as requested during the review phase, we conducted a sensitivity analysis to confirm that minor variations in these hyperparameters do not alter the relative rankings of the evaluated models.

\paragraph{Weight Optimization.} We collected a validation set of 400 complex structural prediction samples. Three human annotators labeled whether the model predictions were semantically equivalent to the ground truth (accounting for acceptable variations where strict string matching fails). Through a grid search, we evaluated different weight configurations based on their alignment accuracy with human annotations.

\paragraph{Sensitivity and Ranking Stability.} We selected four representative weight configurations to test the stability of our benchmark:
\begin{itemize}[leftmargin=1.5em, topsep=2pt, itemsep=0pt]
    \item \textbf{Config A (Optimal):} $\alpha=0.3, \beta=0.2, \gamma=0.1, \delta=0.4$. Achieves the highest human alignment accuracy (\textbf{92\%}).
    \item \textbf{Config B (Equal Weights):} $\alpha=0.25, \beta=0.25, \gamma=0.25, \delta=0.25$. Achieves \textbf{91\%} human alignment.
    \item \textbf{Config C (Lexical-Heavy):} $\alpha=0.4, \beta=0.3, \gamma=0.3, \delta=0.0$. Drops semantic embeddings entirely. Achieves \textbf{88\%} human alignment.
    \item \textbf{Config D (Semantic-Heavy):} $\alpha=0.0, \beta=0.3, \gamma=0.3, \delta=0.4$. Heavily penalizes Levenshtein distance, focusing on semantics and token matching. Achieves \textbf{90\%} human alignment.
\end{itemize}

As shown in Table \ref{tab:sensitivity_analysis}, we re-evaluated five leading models across these distinct configurations. While the absolute Overall F1 scores exhibit minor fluctuations depending on the strictness of the weights, the \textbf{relative ranking of the models remains strictly consistent} (Qwen3-VL-8B (+GRPO) > Qwen3-VL-8B (SFT) > Gemini-3-Pro > Qwen3-VL-Plus > GPT-5) across all scenarios. This empirical proof firmly validates that our evaluation metric is robust, and the superior reasoning and structural capabilities of our Metric-Aware GRPO framework are not artifacts of hyperparameter selection.

\begin{table}[h]
\centering
\small
\resizebox{\columnwidth}{!}{%
\begin{tabular}{l|cccc}
\hline
\multirow{2}{*}{\textbf{Model}} & \multicolumn{4}{c}{\textbf{Overall F1 Score under Different Configs}} \\
 & \textbf{Config A (Opt.)} & \textbf{Config B} & \textbf{Config C} & \textbf{Config D} \\ \hline
GPT-5 & 0.7076 & 0.7015 & 0.6945 & 0.7070 \\
Gemini-3-Pro & 0.7373 & 0.7305 & 0.7306 & 0.7312 \\ 
Qwen3-VL-Plus & 0.7210 & 0.7176 & 0.7170 & 0.7183 \\ \hline
Qwen3-VL-8B (SFT) & 0.7736 & 0.7685 & 0.7672 & 0.7693 \\
Qwen3-VL-8B (+GRPO) & \textbf{0.7950} & \textbf{0.7938} & \textbf{0.7937} & \textbf{0.7945} \\ \hline
\end{tabular}%
}
\caption{Hyperparameter sensitivity analysis on the Overall F1 score. The evaluation metric remains highly stable, with the relative performance rankings strictly preserved regardless of the weight distribution.}
\label{tab:sensitivity_analysis}
\end{table}

\subsection{Qualitative Case Study}
\label{app:case_study}

To visually demonstrate the challenges of \datasetname\ and the effectiveness of our training pipeline, we present a detailed comparison between the \textbf{One-shot Base Model (Qwen3-VL-4B)} and our final \textbf{Fine-tuned Model (Ours, SFT+GRPO)} on a complex hotel receipt.

\begin{figure*}[h]
    \centering
    \includegraphics[width=\textwidth]{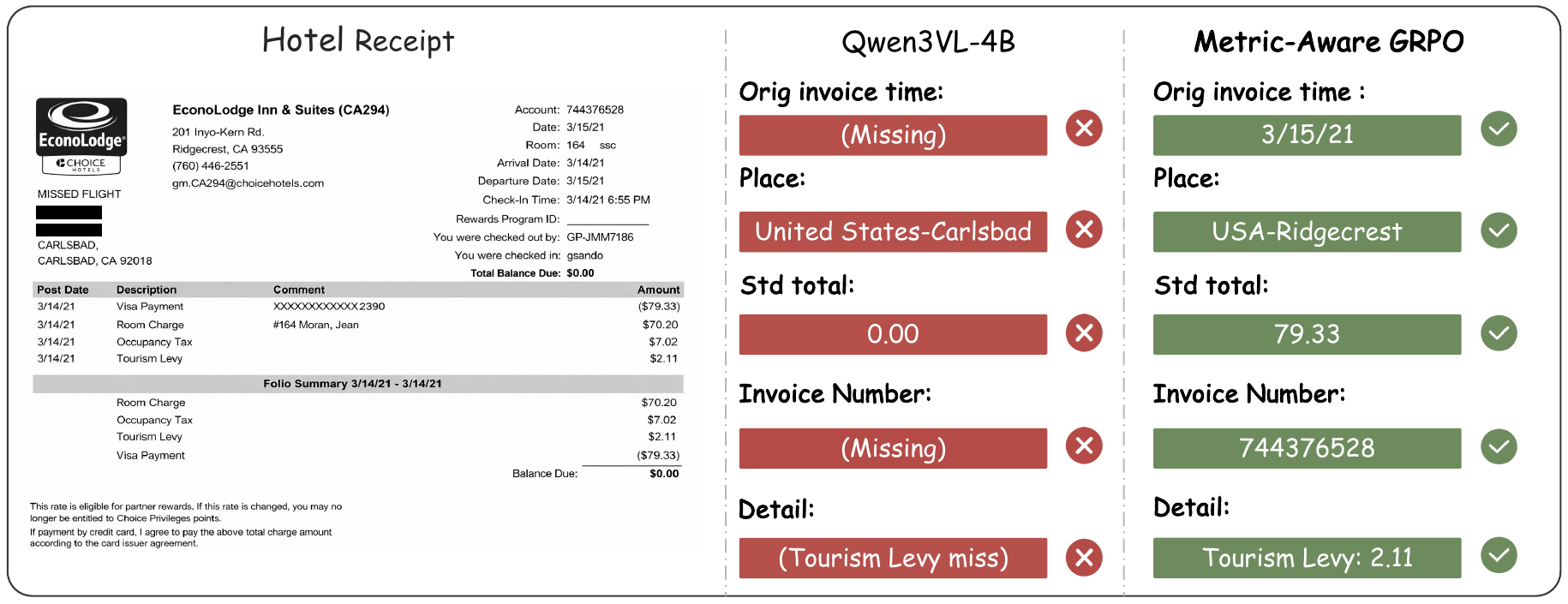} 
    \caption{\textbf{Qualitative comparison on a complex hotel folio.} The \textbf{One-shot Base Model} (middle) falls into common visual and logical traps: extracting the billing address instead of the hotel location, and mistaking the "Balance Due" ($0.00$) for the total amount. In contrast, our \textbf{Fine-tuned Model} (right) correctly infers the semantic roles of fields and adheres to financial logic.}
    \label{fig:case_study}
\end{figure*}

As shown in Figure \ref{fig:case_study}, the input image is a hotel folio from \textit{EconoLodge}. This sample features a scattered layout with multiple address blocks and a "Balance Due" table, posing significant cognitive hurdles. The comparison highlights four key improvements:

\paragraph{1. Spatial Reasoning and Disambiguation (Task 3).}
The document contains two distinct addresses: the hotel's physical address (top, "Ridgecrest") and the customer's billing address (bottom-left, "Carlsbad"). The Base Model creates a hallucination by concatenating "United States" with the distractor address "Carlsbad" for the \texttt{place} field. This is a typical spatial reasoning failure. Our model, aligned via SFT+GRPO, correctly identifies the semantic role of the top address block, accurately extracting "USA-Ridgecrest".

\paragraph{2. The "Balance Due" Trap (Task 2 \& 3).}
For the \texttt{std\_total} field, the Base Model extracts "$0.00$" because the receipt explicitly states "Total Balance Due: \$0.00" (indicating the bill has been paid). This reveals a lack of financial logic in general-purpose models. Our model correctly reasons that the effective transaction amount is the sum of charges (or the payment amount), correctly extracting "$79.33$".

\paragraph{3. Semantic Mapping of Identifiers and Dates (Task 1).}
The receipt does not explicitly label an "Invoice Number" or "Invoice Date" using standard terminology. Instead, it uses the term "Account: 744376528" for the invoice identifier and presents the issuance date under the heading "Date". The Base Model fails to recognize these semantic synonyms, returning \textit{Missing} for both fields. In contrast, our model successfully maps the semantically equivalent "Account" to the target \texttt{invoice\_number} field and "date" to \texttt{orig\_invoice\_date} field, demonstrating robust domain adaptation and semantic understanding.

\paragraph{4. Structural Completeness (Task 4).}
In the \texttt{detail} list extraction, the Base Model misses the last line item ("Tourism Levy"), likely due to its visual separation from the main table body or its small font size. Our model achieves full recall, capturing all line items including the tax details. This structural completeness is crucial for the arithmetic consistency reward used during GRPO training.

In summary, this case illustrates that \textbf{One-shot General MLLMs often fail to distinguish semantic roles (e.g., Service vs. Billing address) and lack domain-specific financial logic (e.g., Total vs. Balance).} Our dataset and training pipeline effectively bridge these gaps.

\end{document}